# A Survey of Some Density Based Clustering Techniques

Rupanka Bhuyan[1], Samarjeet Borah[2]

[1]Department of IT & Mathematics
Icfai University Nagaland,
6th Mile, Sovima, Dimapur–797112
[2]Dept. of Computer Science & Engineering,
Sikkim Manipal Institute of Technology,
Majitar, Rangpo, East Sikkim–737132

[1]rupanka@yahoo.com, [2]samarjeetborah@gmail.com

*Abstract*— Density Based Clustering are a type of Clustering methods using in data mining for extracting previously unknown patterns from data sets. There are a number of density based clustering methods such as DBSCAN, OPTICS, DENCLUE, VDBSCAN, DVBSCAN, DBCLASD and ST-DBSCAN. In this paper, a study of these methods is done along with their characteristics, advantages and disadvantages and most importantly – their applicability to different types of data sets to mine useful and appropriate patterns.

## I. Introduction

Cluster Analysis or Clustering is defined as a method wherein a given set of data objects are grouped into distinctly different sets or groups. Each such set contains objects which are similar to other objects in the same set; consequently objects in different sets are dissimilar to one another.

It is worth mentioning that, from the same data set, different clusterings may be obtained by different clustering methods.

Clustering is generally performed by clustering algorithms using computers on large data sets; essentially it is not possible to perform this manually.

One of the primary motivations for clustering is discovering previously unknown groups inside a data set [2].

Objects within a cluster are "similar" to one another; wherein similarity is calculated or derived in terms of "closeness" i.e. how close two objects are in space. This is done by using specific distance function.

Attributes of a cluster such as – its diameter (the maximum distance between any two objects in the cluster) represents the "quality" of a cluster.

Applications of Cluster Analysis is done in a variety of diverse fields such as knowledge discovery in web searches, business intelligence, image pattern recognition, intrusion detection, intrusion prevention, genomics, speech processing, and fraud detection, to name a few.

### A. Density Based Clusters

Clusters may be looked at as dense regions in the data space; where clusters are separated by a sparse region containing "relatively few" data. Given this assumption, a cluster can either be of "regular" or "arbitrary" shape.

The notion behind density based clustering is to detect clusters of non-spherical or arbitrary shapes.

Some of the common density based clustering techniques are DBSCAN, OPTICS, VDBSCAN, DVBSCAN, DBCLASD, ST-DBSCAN and DENCLUE [1][4] are reviewed in this paper.

## II. Density Based Clustering Methods

### A. Density-Based Spatial Clustering of Applications with Noise (DBSCAN)

One of the earliest density based clustering methods is DBSCAN.

DBSCAN discovers high density regions in spatial databases with noise and creates clusters out of them. [4].

#### 1) Advantages of DBSCAN

- Clusters of arbitrary shape can be detected
- No prior knowledge about the number of clusters is required
- There is a notion of noise (objects not belonging to any cluster)
- Only two (2) input parameters (ε – *radius* and *MinPts* – minimum number of points) and is mostly insensitive to the ordering of the points in the database

#### 2) Disadvantages of DBSCAN

- Proper determination of the initial values of the parameters *ε* and *MinPts* is difficult
- For *n* data objects, without any special structure or spatial indexing, the computational complexity is O($n^2$); with spatial indexing it is O(n log n).

- If there is variation in the density, noise points are not detected

### B. The VDBSCAN (Varied Density Based Spatial Clustering of Applications with Noise)

The VDBSCAN algorithm can detect clusters with varied density. Also, the method automatically selects several values of the input parameter *Eps* for different densities. Even, the parameter *k* is automatically generated based on the characteristics of the datasets [5].

This algorithm consists of two steps, choosing parameters $Eps_i$ and cluster with varied densities. The procedure for this algorithm is as follows,

- It calculates and stores *k*-dist for each project and partition the *k*-dist plots.
- The number of densities is given intuitively by *k*-dist plot.
- The parameter $Eps_i$ is selected automatically for each density.
- Scan the dataset and cluster different densities using corresponding $Eps_i$
- Display the valid cluster with respect to varied density.

*1) Advantages of VDBSCAN*
- Clusters of varied densities can be detected
- Automatic selection of several values of input parameter $Eps_i$ for different densities

*2) Disadvantages of VDBSCAN*
- Time Complexity is high i.e. $O(n^2)$
- For n data objects. If spatial indexing is used, the complexity becomes O(n log n). But, maintaining a spatial index is time consuming.

### C. DVBSCAN (A Density Based Algorithm for discovering Density Varied Clusters in Large Spatial Databases)

This algorithm handles local density variation within the cluster. The input parameters used in this algorithm are minimum objects (μ), radius, and threshold values (α, λ).

The algorithm calculates the growing cluster density mean and then the cluster density variance for any core object, which is supposed to be expanded further by considering density of its *E-neighborhood* with respect to cluster density mean. If cluster density variance for a core object is less than or equal to a threshold value and is also satisfying the cluster similarity index, then it will allow the core object for expansion [6].

*1) Advantages of DVBSCAN*
- It handles local density variation within the cluster
- It outperforms DBSCAN specially in case of local density

*2) Disadvantages of VDBSCAN*

- Time Complexity is high i.e. $O(n^2)$ for a set of *n* data objects

### D. DBCLASD (A Distribution–Based Clustering Algorithm for Mining Large Spatial Databases)

The DBCLASD [7] algorithm detects clusters with arbitrary shape; moreover, it does not require any input parameters. For large spatial databases, the efficiency of DBCLASD is of prime interest.

The algorithm is described as follows:

- It is an incremental algorithm i.e. a point is added to a cluster based only on the points processed so far and without considering the whole database.
- It incrementally augments an initial cluster by its neighbouring points as long as the nearest neighbour distance of the resulting cluster fits the expected distance distribution. A set of candidates of a cluster is constructed using region queries which are supported by Spatial Access Methods (SAM). The calculation of m is based on the model of uniformly distributed points inside the cluster *C*. Let *A* be the area of *C* and *N* be the number of its elements. A necessary condition for *m* is

    $N \times P(NNdist_C(P) > m) < 1$

    When inserting a new point *p* into cluster *C*, a circle query with center *P* and radius m is performed and the resulting points are considered as new candidates.
- The incremental approach implies an inherent dependency of the discovering clusters from the order of generating and testing candidates. The order of testing the candidates is crucial. Candidates which are not accepted by the test for the first time are called unsuccessful candidates. To minimize the dependency on order of testing, the following two features are considered:
    i. Unsuccessful candidates are not discarded but they are tried again later.
    ii. Points already assigned to some cluster may switch to another cluster later.

The testing of candidates are performed in two steps are as follows,

i. The current cluster is augmented by the candidate
ii. Chi-square test is used to verify the hypothesis that the nearest neighbour distance set of the augmented cluster still fits the expected distance distribution.

*1) Advantages of DBCLASD*
- It does not require any input parameter
- It is suitable for clusters with uniformly distributed points

### E. ST-DBSCAN (Spatial–Temporal Density Based Clustering)

The ST-DBSCAN algorithm is suitable for applications involving spatial – temporal data such as Geographical Information Systems, Medical Imaging, and Weather Forecasting. [8]

The technique starts with the first point *p* in database *D*.

i. This point *p* is processed according to DBSCAN algorithm and next point is taken.
ii. Retrieve_Neighbors (object, $Ep_1$, $Ep_2$) function retrieves all objects density-reachable from the selected object with respect to *Eps1*, *Eps2* and *MinPts*. If the returned points in *Eps-neighborhood* are smaller than *MinPts* input, the object is assigned as noise.
iii. The points marked as noise can be changed later i.e. the points are not directly density-reachable but they will be density reachable.
iv. If the selected point is a core object, then a new cluster is constructed. Then all the directly-density reachable neighbours of this core objects is also included.
v. Then the algorithm iteratively collects density reachable objects from the core object using stack.
vi. If the object is not marked as noise or it is not in a cluster and the difference between the average value of the cluster and new value is smaller than *ΔE*, it is placed into the current cluster.
vii. If two clusters *C1* and *C2* are very close to each other, a point *p* may belong to both *C1* and C2. Then point *p* is assigned to the cluster which is discovered first.

*1) Advantages of ST-DBSCAN*
➤ Clustering of temporal–spatial data can be performed based on non–spatial, spatial and temporal attributes
➤ It can detect noise points even when it is of varied density by assigning density factor to each cluster

*F. OPTICS (Ordering Points to Identify the Clustering Structure) Algorithm*

To overcome the difficulty in using one set of global parameters in clustering analysis, a cluster analysis method called OPTICS was proposed [4].

It is an indirect method i.e. OPTICS does not explicitly produce a data set clustering; instead, it outputs a cluster ordering. This is a linear list of all objects under analysis and represents the *density-based clustering structure* of the data. Objects in a denser cluster are listed closer to each other in the cluster ordering. This ordering is equivalent to density-based clustering obtained from a wide range of parameter settings.

*1) Advantages of OPTICS*
➤ It does not require the user to provide any density threshold

*2) Disadvantages of OPTICS*
➤ The Time Complexity is quite high i.e. $O(n^2)$ for a database of *n* objects. For a spatially indexed set, the complexity is O(n log n).

*G. The DENCLUE (DENsity-based CLUstEring) Algorithm*

The DENCLUE algorithm is based on a set of density distribution functions. It is an improvement over the basic density based clustering methods DBSCAN and OPTICS in terms of density estimation.

The technique is briefly described below:
i. A statistical density estimation method is used for estimating the *kernel density*. This results in the *local density maxima value.*
ii. Clusters can be formed from this *local density maxima value.*
iii. If the *local density value* is very small, then the objects of clusters are discarded as *noise*

In this method, objects under consideration are added to a cluster through density attractors using a step wise hill-climbing procedure.

*Advantages of OPTICS*
➤ Sensitivity of density (owing to the radius parameter ε) is removed
➤ Clusters of arbitrary shape can be found
➤ This method is invariant against noise

III. SUMMARIZED COMPARISON OF THE METHODS

Given in Fig. 1 is a table summarizing the characteristics of the various density based clustering methods under consideration.

IV. CONCLUSION

Clustering is an emerging and promising area of study nowadays. Its application in varied fields of data mining for previously undiscovered subtle patterns has given this study a prime importance.

In this survey, a group of Density Based Clustering methods have been analysed. It has been found that DBSCAN is a very basic and one of the earliest methods which can be used on data sets with noise. The OPTICS algorithm on the other hand can overcome the shortcoming of DBSCAN which does not yield accurate results for data with variable densities. DENCLUE on the other hand can deal with the inherent density sensitivity problem of both the DBSCAN and OPTICS, which arises due to the estimation of the radius *ε*.

| Algorithm | Varied Density | Primary Input Requirement | Time Complexity | Cluster Type | Type of Data |
|---|---|---|---|---|---|
| DBSCAN | No | Cluster Radius, Minimum number of Objects | $O(n^2)$; $O(nLogn)$ for spatial indexed data | Arbitrary Shaped | Spatial Data with Noise |
| VDBSCAN | Yes | Automatically Generated | $O(n^2)$ | Arbitrary Shaped | Spatial Data with Varied Density |
| DVBSCAN | Yes | Two Input Parameters to be given by User | $O(n^2)$ | Arbitrary Shaped | Spatial Data with Varied Density |
| DBCLASD | Yes | Automatically Generated | $O(n^2)$ | Arbitrary Shaped | Spatial Data with Uniformly Distributed Points |
| ST-DBSCAN | No | Three Input Parameters to be given by User | $O(n^2)$ | Arbitrary Shaped | Spatio-Temporal |
| OPTICS | Yes | Density Threshold | $O(n^2)$; $O(nLogn)$ for spatial indexed data | Arbitrary Shaped | Spatial Data with Varied Density |
| DENCLUE | Yes | Radius | $O(n^2)$ | Arbitrary Shaped | Spatial Data with Varied Density |

Fig.1: Comparision of various Density Based Clustering Algorithms

Other variants of the DBSCAN algorithm are also studied. For instance, the VDBSCAN method can detect clusters with varied densities. The DVBSCAN algorithm is particularly applicable in situations where *clusters of varied density* are to be formed in *extremely large spatial databases*. Again, for very large spatial databases with *uniformly distributed points* the DBCLASD is of primary importance. Clustering on Temporal – Spatial data can be carried out by using the ST-DBSCAN which assigns specific density factors to the clusters for detecting noise points.

As it can be seen from this study, the applicability and appropriateness of a specific method is mandated by the type of the data set.

Given below is a summarized table containing a comparison of all the algorithms under consideration.